# Combining ECG Foundation Model and XGBoost to Predict In-Hospital Malignant Ventricular Arrhythmias in AMI Patients

Shun Huang[1], Wenlu Xing[2], Shijia Geng[3], Hailong Wang[4], Guangkun Nie[1,5], Gongzheng Tang[1,5], Chenyang He[1], Shenda Hong[1,5,*]

1. National Institute of Health Data Science, Peking University, Beijing, China.
2. Fuwai Central China Cardiovascular Hospital, Zhengzhou, China.
3. HeartVoice Medical Technology, Hefei, China.
4. Department of Physiology and Pathophysiology, School of Basic Medical Sciences, State Key Laboratory of Vascular Homeostasis and Remodeling, Peking University, Beijing, China.
5. Institute of Medical Technology, Peking University Health Science Center, Beijing, China.

*Correspondence: Shenda Hong, hongshenda@pku.edu.cn

## Abstract

**Background and Objective:** Malignant ventricular arrhythmias (including ventricular tachycardia, VT, and ventricular fibrillation, VF) following acute myocardial infarction (AMI) are major causes of in-hospital sudden cardiac death. Early identification of high-risk patients remains a significant clinical challenge. Traditional risk scoring systems have limited predictive performance, while end-to-end deep learning models often lack interpretability, which constrains their clinical application. This study aimed to develop a hybrid predictive framework that integrates a large-scale electrocardiogram (ECG) foundation model (ECGFounder) with an interpretable machine learning algorithm (XGBoost) to improve both the accuracy and interpretability of VT/VF risk prediction in AMI patients.

**Methods:** A total of 6,634 ECG recordings from AMI patients admitted to Fuwai Central China Cardiovascular Hospital between 2019 and 2023 were analyzed, among which 175 patients experienced in-hospital VT/VF events. The ECG data were first processed by the ECGFounder model to extract 150-dimensional diagnostic probability features. These features were then refined through feature selection and hyperparameter tuning before being used to train and test an XGBoost classifier. Model performance was evaluated using AUC, F1-score, precision, and recall metrics. The SHAP method was employed to interpret model predictions and assess feature importance.

**Results:** The ECGFounder + XGBoost hybrid model achieved an AUC of 0.801 and an F1-score of 0.391 on the test set, outperforming the KNN model (AUC 0.677), the RNN model (AUC 0.676), and the end-to-end 1D-CNN model (AUC 0.720). SHAP analysis revealed that the model-identified key features were highly consistent with clinical knowledge, where "normal sinus rhythm" and "early repolarization" acted as protective factors, whereas "premature ventricular complexes" and "incomplete left bundle branch block" were strong risk predictors.

**Conclusion:** The proposed hybrid framework demonstrated both high predictive performance and clinical interpretability, providing a novel paradigm for in-hospital VT/VF risk prediction in AMI patients. This study validates the feasibility of using foundation model outputs as automated feature engineering and offers a promising direction for building trustworthy, explainable AI-based clinical decision support systems.

## Introduction

Cardiovascular disease (CVD) represents a major and ongoing global health crisis. According to the World Health Organization, CVD remains the leading cause of death worldwide[1]. Approximately 85% of CVD-related deaths result from acute myocardial infarction (AMI) and stroke[1,2]. Despite substantial improvements in AMI incidence and mortality over recent decades due to advances in reperfusion therapy and guideline-directed medical management, a considerable proportion of patients continue to experience severe in-hospital complications[3,4]. Among these, malignant ventricular arrhythmias—including ventricular tachycardia (VT) and ventricular fibrillation (VF)—remain one of the primary causes of sudden cardiac death, contributing to persistently high short-term mortality in this population[5,6]. Early identification of patients at high risk for such arrhythmic events is a critical yet unresolved clinical challenge, largely due to the transient and unpredictable nature of their onset[7].

To address this unmet need, numerous approaches have been proposed in both academic and clinical settings. Conventional risk stratification tools such as the GRACE and TIMI risk scores have demonstrated certain prognostic value, but their predictive performance remains limited[8,9]. Similarly, traditional ECG-based indicators such as T-wave alternans and QT dispersion are difficult to compute and have limited predictive strength when used in isolation[10,11]. In recent years, deep learning has shown remarkable potential in cardiovascular diagnostics, as it can automatically learn complex features directly from raw ECG signals[12-14]. Deep learning–based models have been developed to predict the occurrence of malignant ventricular arrhythmias using ECG data[15,16]. However, most end-to-end deep learning models operate as black boxes, with limited interpretability that constrains clinical acceptance and trust[17]. In contrast, traditional machine learning models offer greater interpretability but often depend heavily on complex, manually designed feature engineering that requires substantial domain expertise[18].

To bridge the gap between predictive performance and interpretability, this study explores a hybrid predictive framework that combines a foundation model with a traditional machine learning algorithm. Foundation models represent a new paradigm in medical artificial intelligence[19, 20], referring to large-scale AI models pretrained on vast and diverse datasets that have achieved outstanding performance across various clinical diagnostic tasks[21]. Our previously developed ECG foundation model, ECGFounder, exemplifies this approach. Trained on a dataset of over 10 million ECGs covering 150 diagnostic categories, ECGFounder has demonstrated strong representation capability for ECG signals[14].

The central hypothesis of this study is that the 150-dimensional diagnostic probability outputs generated by ECGFounder can serve as a form of automated, high-level feature engineering. To validate this hypothesis, we collected admission ECG data from 6,634 AMI patients at Fuwai Central China Cardiovascular Hospital, and investigated whether these diagnostically rich feature vectors could effectively train a traditional machine learning model to achieve accurate in-hospital prediction of VT/VF events.

In summary, this study aims to establish a novel hybrid predictive framework that integrates a large-scale ECG foundation model with XGBoost, providing a new perspective for post-AMI risk stratification. Furthermore, it seeks to validate an innovative feature engineering paradigm, demonstrating that the diagnostic outputs of a foundation model can serve as effective features for downstream predictive tasks. Importantly, this approach aims to balance predictive performance with clinical interpretability, identifying ECG patterns highly associated with malignant ventricular arrhythmias and thereby offering more transparent support for clinical decision-making.

## Methods

### Study Population and Data Matching

The study data were obtained from Fuwai Central China Cardiovascular Hospital and included electrocardiogram (ECG) records of patients discharged between December 2019 and December 2023 with a primary diagnosis of acute myocardial infarction (AMI). These records formed the original ECG dataset used in this project. The exclusion criteria were as follows: 1.Age younger than 18 years; 2. History of ventricular tachycardia (VT) or ventricular fibrillation (VF) before hospital admission; 3. Absence of ECG data prior to the onset of in-hospital VT or VF; 4.Missing in-hospital ECG data.

To extract ECG data corresponding to the target population, a multi-step matching process was implemented. In the first stage, patient identification numbers from the curated AMI cohort were cross-referenced with the ECG database to identify all ECG records corresponding to the selected patients. Since patients might have multiple ECGs recorded at different time points, including during previous or subsequent hospital visits, a second-stage matching was performed. In this step, ECG records were filtered based on each patient's admission and discharge dates, retaining only those recorded during the hospitalization period for AMI. Because the number of ECGs collected per patient varied substantially, a third-stage screening was conducted to minimize inter-individual variability and to better reflect real-world clinical conditions. In this third step, only the first ECG acquired after hospital admission was retained as the representative sample for each patient in subsequent analyses.

All matched ECG data were then categorized into two groups. The positive group consisted of patients who developed VT, VF, or both during hospitalization. The negative group included patients who did not experience any malignant ventricular arrhythmias during their hospital stay. Both groups were randomly divided into training (80%) and testing (20%) subsets using stratified sampling to ensure that the ratio of positive to negative cases remained consistent with that of the original dataset.

**ECGFounder Base Model and Feature Extraction**

The ECGFounder model is built upon the RegNet framework and adopts a staged network architecture that integrates multi-level convolutional and attention mechanisms to effectively capture both spatial and temporal features from large-scale electrocardiogram (ECG) data. The model was trained on the HEEDB dataset, which is currently one of the largest known ECG databases. HEEDB contains more than ten million ECG recordings collected from over one million patients and includes up to 150 diagnostic label categories, enabling comprehensive analysis and diagnosis of various cardiovascular diseases[14].

In this study, these 150 diagnostic categories were used as a structured feature system to support the prediction of more complex cardiovascular outcomes. The original ECG recordings were input into the ECGFounder model, which generated for each ECG a set of 150 predicted probabilities corresponding to these diagnostic labels. These probability values, expressed as percentages, were then used as feature vectors for subsequent downstream modeling and training tasks.

**Feature Engineering and Model Training**

To improve the generalization ability of the model and reduce the risk of overfitting, we implemented a feature selection pipeline consisting of three key steps: (1) removal of low-variance features, identifying and deleting those features that had the same value in more than 90% of samples; (2) handling of highly correlated features, where pairs of features with a correlation coefficient greater than 0.70 were examined and one feature in each pair was removed; (3) entropy-based ranking, in which all features were ranked according to their information entropy, and features were incrementally added in descending order of entropy to construct the model, selecting the subset with the highest AUC for subsequent model training. The XGBoost classifier was used for model training. To achieve optimal performance, hyperparameter tuning was performed through grid search by testing predefined combinations of parameters such as the number of trees, maximum tree depth, and learning rate, combined with stratified k-fold cross-validation to identify the best-performing configuration on the training set. The average precision score was chosen as the evaluation metric for hyperparameter optimization, as it provides a more comprehensive reflection of the model's ability to identify minority classes compared with accuracy in imbalanced datasets.

The final XGBoost model was trained on the full training set using the optimized hyperparameters. The model performance was then evaluated on an independent test set using several metrics, including accuracy, precision, recall, F1-score, area under the receiver operating characteristic curve (AUC-ROC), and area under the precision–recall curve (AUC-PRC).

**Model Comparison**

To identify the downstream model with the best performance, we trained and evaluated KNN and RNN models using the same procedure and compared their results with those of the XGBoost model. These three models represent different categories of machine learning algorithms. KNN is a traditional non-parametric, distance-based algorithm that performs classification by measuring the similarity between samples[22]. In contrast, the RNN model represents a sequential learning approach capable of capturing complex nonlinear relationships and potential temporal dependencies in data[23].

To demonstrate the performance advantage of combining ECGFounder with downstream models, we conducted a comparative experiment by directly training the base architecture of ECGFounder, a one-dimensional convolutional neural network (1D-CNN), on the same dataset and comparing its performance with that of the hybrid ECGFounder-based models.

**Feature Evaluation**

A core objective of this study was to establish a new feature system. The contribution and interpretability of each feature to the model's prediction were considered essential. Therefore, the trained model was analyzed using the SHAP method to quantify each feature's contribution to model prediction, providing both global and local interpretability.

**Error Analysis**

False-negative (FN) samples were selected from the test set and compared with true-positive (TP) samples. For all trained features, independent-sample t-tests were performed to examine whether their value distributions differed significantly between the FN and TP groups. For false-positive (FP) samples, due to their limited number, an in-depth case analysis using SHAP was conducted to observe and evaluate how individual features contributed to positive or negative predictions.

## Results

**Study Population and Cohort Composition**

A total of 10,635 patients who were discharged from Fuwai Central China Cardiovascular Hospital between December 2019 and December 2023 with a primary diagnosis of acute myocardial infarction (AMI) were initially identified from the hospital database, yielding 11,352 ECG recordings. After multiple rounds of matching and screening, a total of 6,634 patients with their first admission ECG were finally included in the analysis. The detailed data selection process is illustrated in Figure 1.

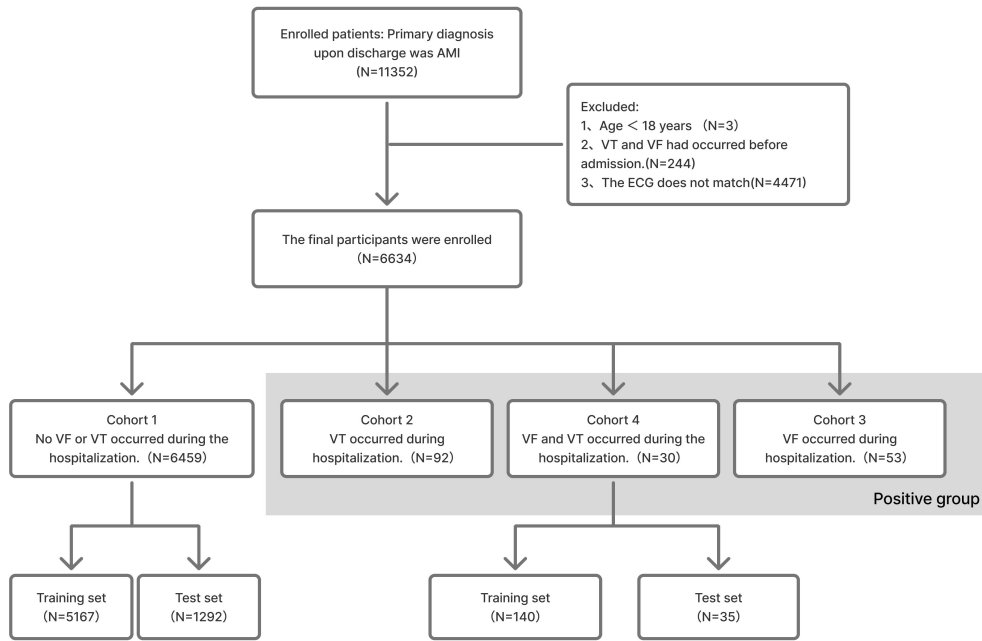

Figure 1 Flowchart of the Study Population Selection Process

Among the final cohort, 175 patients (2.64%) experienced ventricular tachycardia (VT) or ventricular fibrillation (VF) during hospitalization, comprising the positive group. Specifically, 92 patients developed VT, 53 patients developed VF, and 30 patients experienced both events. The remaining 6,459 patients (97.36%) did not experience VT or VF and were classified as the negative group. The demographic and clinical characteristics of the study population are summarized in Table 1.

Table 1 : Demographic Characteristics of the Study Population

| Variables | Mean | Median | Max | Min |
|---|---|---|---|---|
| Age | 62.39 | 62 | 96 | 21 |
| Height | 165.11 | 166 | 190 | 145 |
| Wight | 70.71 | 70 | 183 | 34.5 |
| Gender | Number | Percentage | - | - |
| Male | 4915 | 74.1% | - | - |
| Female | 1719 | 25.9% | - | - |
| Overall | 6634 | 100% | - | - |

**Feature Selection**

Variance analysis was performed on the 150 features output by the ECGFounder model, and no low-variance features were identified. This indicated that the model-generated prediction values exhibited sufficient variability to serve as effective features. Pairwise Pearson correlation coefficients were then calculated among all features, and 32 pairs were found to have correlation coefficients greater than 0.70. One feature from each highly correlated pair was removed, including features 2, 3, 19, 21, 27, 40, 41, 46, 51, 54, 55, 64, 75, 76, 83, 90, 92, 97, 98, 100, 111, 115, 119, 123, 125, 130, 132, 134, 138, 141, and 145.

The remaining 118 features were then ranked in descending order according to their information entropy, and models were sequentially constructed by progressively adding features. Figure 2 illustrates the change in model performance as the number of features increased. When the number of features reached 116, the model achieved the highest AUC value of 0.8014. During this process, features 127 and 126 were excluded. The final set of 116 selected features was used for subsequent model training and analysis.

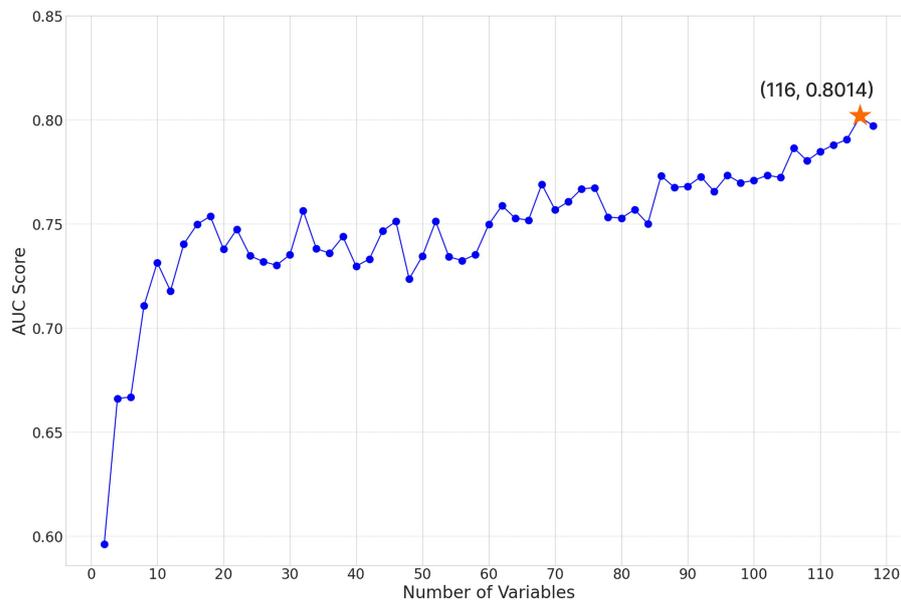

Figure 2 Feature Selection Process Based on Recursive Feature Elimination

**Model Development and Comparison**

In this study, the positive and negative samples were divided into training and testing sets in an 80% to 20% ratio. Among the positive samples, 140 cases were assigned to the training set and 35 cases to the testing set, while among the negative samples, 5,167 were used for training and 1,292 for testing. The 116 selected features were then input into three downstream models, namely XGBoost, KNN, and RNN, to construct classification models. In addition, a 1D convolutional neural network (1D-CNN) model was trained directly on the raw ECG data. The comparison of AUC curves for the four models is shown in Figure 3.

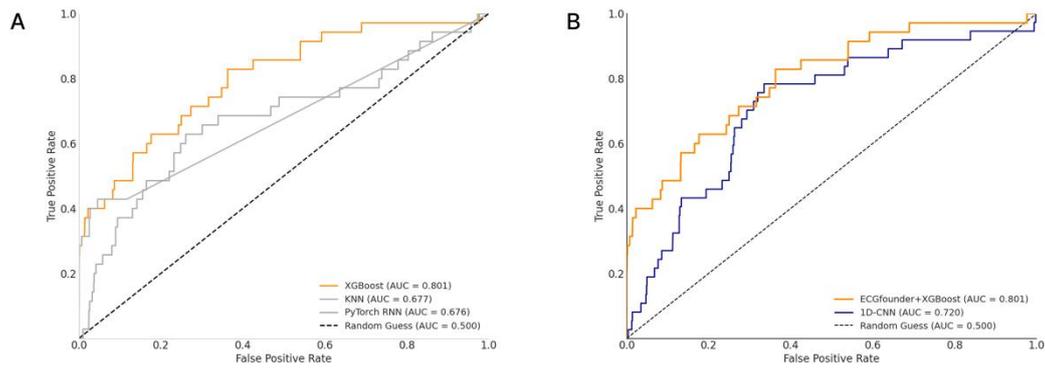

Figure 3 : Comparison of Receiver Operating Characteristic (ROC) Curves for Different Prediction Models.(A) Comparison of downstream models: XGBoost vs. KNN and RNN.(B) Comparison of the composite model (ECGFounder + XGBoost) vs. the 1D-CNN model.

Table 2 presents the comprehensive performance metrics of the four models. Specifically, the ECGFounder + XGBoost hybrid model demonstrated the best overall performance, achieving an AUC of 0.801 and an F1-score of 0.391. The ECGFounder + KNN model achieved an AUC of 0.677 and an F1-score of 0.400; although its F1-score was slightly higher, its overall classification performance remained inferior to that of XGBoost. In contrast, the ECGFounder + RNN model reached an AUC of 0.676 but had an F1-score of 0, indicating ineffective classification. Similarly, the 1D-CNN model achieved an AUC of 0.720 but also had an F1-score of 0.

Table 2 Performance Comparison of Different Prediction Models on the Test Set

| Models | AUC | F1-score | ACC | PRE | SEN | PRC |
|---|---|---|---|---|---|---|
| ECGFounder+Xgboost | 0.801 | 0.391 | 0.979 | 0.818 | 0.257 | 0.368 |
| ECGFounder+KNN | 0.677 | 0.4 | 0.979 | 0.9 | 0.257 | 0.306 |
| ECGFounder+RNN | 0.676 | 0 | 0.974 | 0 | 0 | 0.07 |
| 1DCNN | 0.72 | 0 | 0.974 | 0 | 0 | 0.08 |

To further evaluate model performance, a confusion matrix analysis was conducted for the ECGFounder + XGBoost model. The results showed 1,290 true negatives (TN), 2 false positives (FP), 10 true positives (TP), and 25 false negatives (FN). The model performed well in identifying negative samples but showed relatively limited sensitivity in detecting positive cases.

**Feature Importance Analysis**

To identify which ECG patterns contributed most to the prediction of in-hospital malignant ventricular arrhythmias, we conducted an interpretability analysis of the best-performing XGBoost model using the SHAP method. Figure 4A presents the SHAP bar plot showing the top 20 features ranked by global importance, with the corresponding ECG diagnoses listed in Table 3. The results indicated that the SHAP values of the top 20 features were mostly above or close to 0.1, suggesting that the model was highly sensitive to multiple features.

Table 3 The Top 20 Important Features in the XGBoost Model and Their Clinical Meanings

| Feature | Meaning | Feature | Meaning |
| --- | --- | --- | --- |
| 99 | EARLY REPOLARIZATION | 34 | WITH SINUS ARRHYTHMIA |
| 77 | T WAVE AMPLITUDE HAS INCREASED | 62 | INCOMPLETE LEFT BUNDLE BRANCH BLOCK |
| 32 | ATRIAL FLUTTER | 1 | NORMAL SINUS RHYTHM |
| 147 | WITH 2:1 AV CONDUCTION | 11 | RIGHT BUNDLE BRANCH BLOCK |
| 9 | PREMATURE VENTRICULAR COMPLEXES | 122 | NO P-WAVES FOUND |
| 112 | NONSPECIFIC INTRAVENTRICULAR BLOCK | 108 | WITH VENTRICULAR ESCAPE COMPLEXES |
| 47 | NON-SPECIFIC CHANGE IN ST SEGMENT | 17 | ANTERIOR INFARCT |

The SHAP summary plot in Figure 4B not only illustrated the ranking of feature importance but also revealed the direction of each feature's influence on the predicted risk. Among them, Feature99, Feature77, and Feature1 showed clear negative correlations with the predicted outcome, while Feature62, Feature11, Feature9, Feature112, Feature47, Feature17, Feature101, Feature143, and Feature48 exhibited strong positive associations.

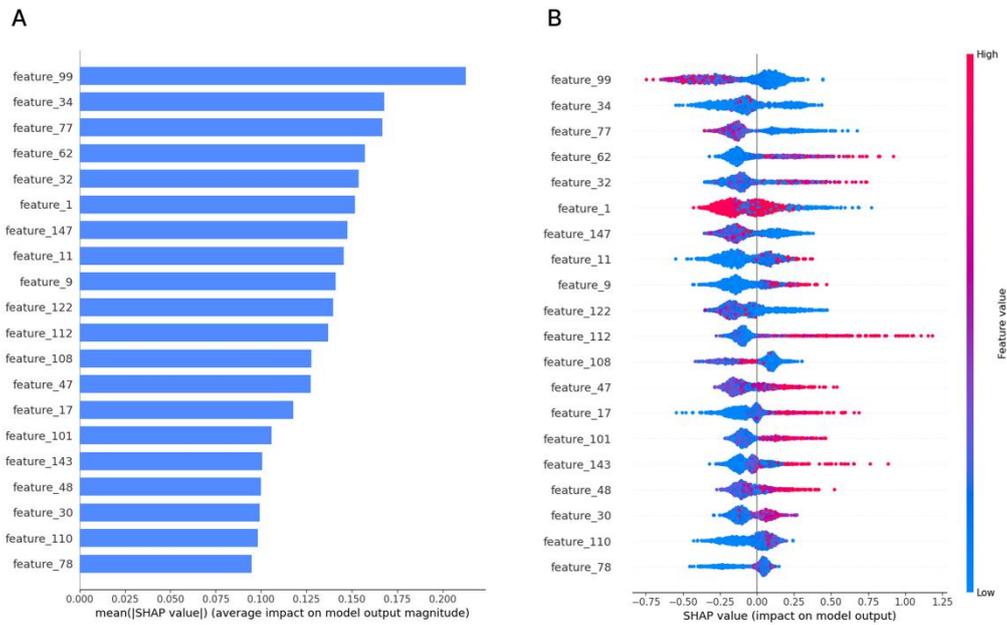

Figure 4  Global Feature Importance Analysis of the XGBoost Model. (A) Bar Plot of the Mean Absolute SHAP Values for the Top 20 Features. (B) SHAP Summary Plot for the Top 20 Features

Figure 5 shows the SHAP scatter plots of the top nine features, providing a more detailed explanation of how each feature influences model predictions across different value ranges. The results further confirmed that Feature99, Feature77, and Feature1 had negative correlations, while Feature62, Feature11, and Feature9 showed positive correlations. Some features, however, displayed nonlinear relationships or potential interaction effects with other features.

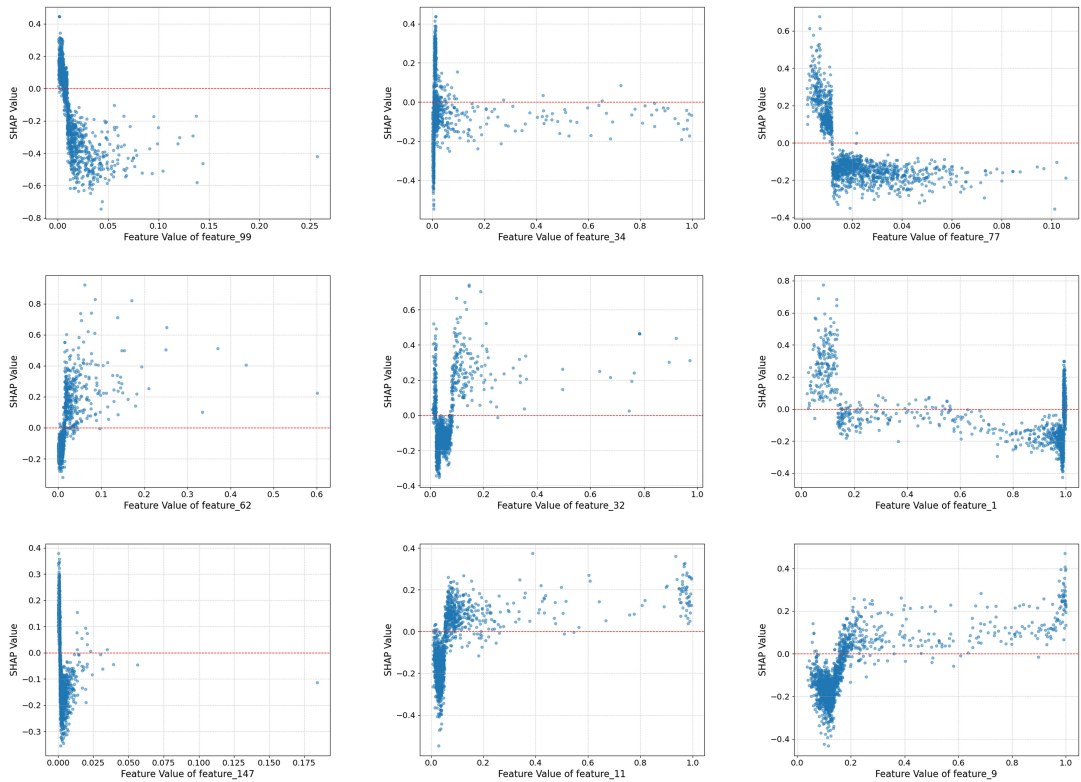

Figure 5　SHAP Scatter Plots for the Top Nine Predictive Features (including Features 99, 34, 77, 62, 32, 1, 147, 11, and 9)

**Error Analysis**

We further analyzed the misclassified samples of the model to investigate its potential limitations.

In the analysis of false-negative (FN) samples, we compared the value distributions of 150 input features between the FN group and the true-positive (TP) group. As shown in Figure 6A, independent-sample t-tests revealed significant differences ($P < 0.05$) between the two groups for several features, including Feature80 (QT HAS LENGTHENED), Feature53 (T WAVE INVERSION LESS EVIDENT), Feature72 (WITH A COMPETING JUNCTIONAL PACEMAKER), and Feature32 (ATRIAL FLUTTER). Specifically, the mean value of Feature80 in the FN group was 0.1038 compared with 0.0205 in the TP group; Feature53 had a mean of 0.0764 in the FN group and 0.1194 in the TP group; Feature72 had a mean of 0.0637 in the FN group and 0.0118 in the TP group; and Feature32 had a mean of 0.2033 in the FN group and 0.0645 in the TP group.

For false-positive (FP) samples, due to their limited number, a detailed case-level analysis was performed. Figure 6B shows a representative FP case in which the patient did not actually experience VT/VF but was incorrectly predicted as high risk by the model. The SHAP waterfall plot revealed that Feature112 (NONSPECIFIC INTRAVENTRICULAR BLOCK), Feature148 (WITH AV DISSOCIATION), and Feature9 (PREMATURE VENTRICULAR COMPLEXES) made strong positive contributions, which ultimately dominated the model's incorrect prediction.

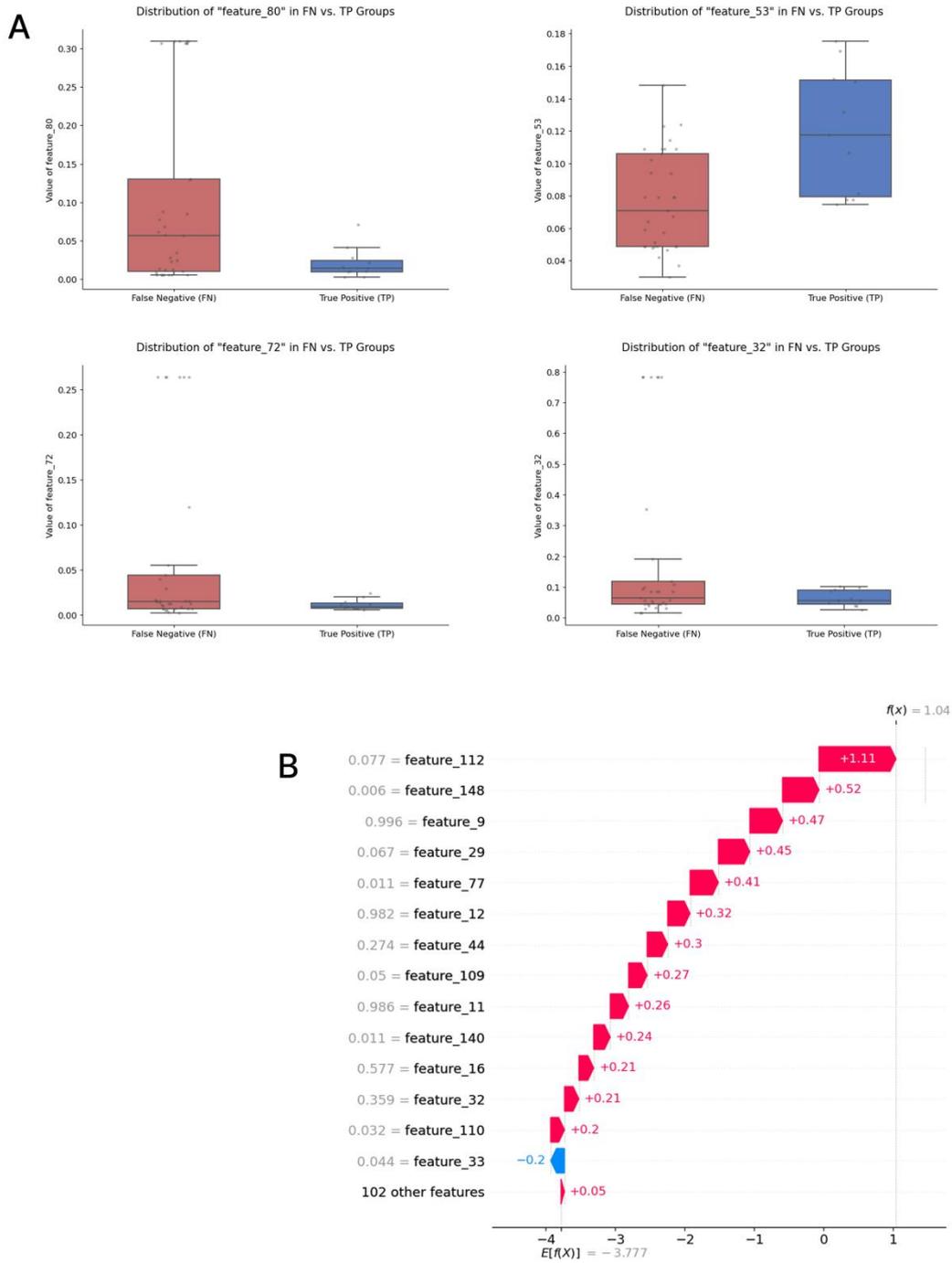

Figure 6 Model Error Analysis (A) Comparison of Key Feature Distributions in False Negative (FN) and True Positive (TP) Samples. (B) SHAP Waterfall Plot Analysis of a Typical False Positive (FP) Case

# Discussion

This study successfully constructed and validated a novel composite predictive framework by integrating a large-scale ECG foundation model (ECGFounder) with an interpretable XGBoost model to effectively predict in-hospital malignant ventricular arrhythmias (VT/VF) in patients with AMI. The results not only confirm the framework's high performance potential but also provide a transparent basis for clinical decision-making through interpretability analysis, effectively bridging the gap between the performance of deep learning models and the need for clinical explainability.

**Main Findings and Clinical Significance**

Our core finding is that the 150-dimensional diagnostic probability vector output by the ECGFounder foundation model serves as a highly efficient, automated form of "feature engineering," providing rich and clinically relevant information for downstream machine learning models. In our tests, the ECGFounder + XGBoost model achieved an AUC of 0.801, outperforming other downstream models as well as an end-to-end 1D-CNN model trained directly on raw ECG data. The F1-scores of 0 for the 1D-CNN and RNN models are particularly noteworthy, indicating that training deep models from scratch on a dataset of this scale and with extreme class imbalance struggles to effectively identify positive samples from the minority class. In contrast, ECGFounder, leveraging its powerful representation capabilities pre-trained on tens of millions of records, successfully extracted key information. This enabled the relatively simple XGBoost model to achieve excellent performance, highlighting the immense potential of foundation models in handling domain-specific, small-sample tasks.

In the feature importance analysis, the key predictors identified by the model were highly consistent with clinical knowledge. For instance, "Normal sinus rhythm" (Feature 1) and "Early repolarization" (Feature 99) were identified as strong protective factors (negative correlation). Conversely, "Incomplete left bundle branch block" (Feature 62), "Right bundle branch block" (Feature 11), and "Ventricular premature beat" (Feature 9) were identified as high-risk predictors (positive correlation). These findings align perfectly with clinical practice, where intraventricular conduction blocks and ventricular ectopy are often considered markers of myocardial instability and potential triggers for arrhythmic events. This not only validates the model's clinical efficacy but also provides a solid foundation for clinicians to understand and trust its predictions.

Furthermore, SHAP analysis revealed that the top 20 features all made significant contributions to the model's decisions, with their mean absolute SHAP values approaching or exceeding 0.1. This indicates that the model's decision-making is multi-dimensional and robust, not reliant on a few sporadic, strong features. Instead, it successfully learned a complex decision rule governed by the interplay of multiple clinically relevant indicators, which also demonstrates the effectiveness of the feature engineering process.

**Model Limitations and Error Analysis**

Despite the model's overall strong performance, an in-depth analysis of its performance and errors revealed several key issues. The confusion matrix shows a clear asymmetry in the types of errors. On one hand, the model produced 25 false negatives (FN) compared to only 10 true positives (TP), leading to a low sensitivity and implying a high risk of missed diagnoses. On the other hand, the model generated only 2 false positives (FP), demonstrating extremely high precision. This characteristic of low sensitivity and high precision suggests that the model can serve as a high-specificity rule-out tool. For the vast majority of patients predicted as negative, they can be confidently identified as low-risk. Conversely, for the few patients predicted as positive, the result should be regarded as a high-confidence warning signal, necessitating further in-depth confirmation.

Further error analysis revealed that the high number of false negatives (missed diagnoses) may be attributable to the model's insufficient weighting of key risk patterns such as "QT prolongation" (Feature 80) and "Atrial flutter" (Feature 32), thus failing to capture some high-risk patients. Meanwhile, the scarcity of false positive cases suggests that the model's decision boundary is relatively strict, with misjudgments occurring only in rare instances dominated by the extremely high contribution of strong risk features like "Nonspecific intraventricular block."

**Strengths and Limitations of the Study**

The primary advantage of the proposed composite framework lies in its success at bridging the gap between traditional clinical risk scores and end-to-end deep learning models. Unlike scores such as GRACE and TIMI, which rely on a limited set of clinical variables, our framework can directly mine deep information from the ECG, one of the most fundamental and data-rich sources of examination data in cardiovascular medicine. Concurrently, compared to deep learning models that lack transparency, our approach uses SHAP analysis to anchor the basis of its predictions to specific ECG diagnoses understandable to clinicians, thereby greatly enhancing the model's trustworthiness and potential for clinical application. This methodological innovation, along with the in-depth exploration of model interpretability, constitutes the core value of this research.

Nevertheless, this study has several limitations. First, as a single-center, retrospective study, the external validity of its conclusions requires further verification through multi-center, prospective research. Second, the extremely low prevalence of positive samples in the dataset (2.64%) directly resulted in a trade-off where high specificity was achieved at the cost of a relatively modest sensitivity (recall) of 0.257. Third, this study utilized only the initial ECG upon admission, failing to capture the dynamic evolution of ECGs during hospitalization. Finally, the model is currently based solely on ECG features; integrating other dimensions of clinical data (such as electrolyte levels, ejection fraction, and medication information) could further enhance its predictive performance.

## Conclusion

In summary, this study successfully developed and validated a composite framework combining an ECG foundation model with XGBoost for predicting the risk of in-hospital malignant ventricular arrhythmias in AMI patients. This approach not only demonstrated excellent predictive performance but also, through in-depth interpretability analysis, provided a transparent basis for clinical decision-making, striking a meaningful balance between model performance and clinical utility. This work offers a successful paradigm for leveraging large-scale foundation models to empower clinical prediction tasks and represents an important direction in building the next generation of intelligent and trustworthy clinical decision support tools.

(1) Roth, G. A.; Johnson, C.; Abajobir, A.; Abd-Allah, F.; Abera, S. F.; Abyu, G.; Ahmed, M.; Aksut, B.; Alam, T.; Alam, K.; et al. Global, Regional, and National Burden of Cardiovascular Diseases for 10 Causes, 1990 to 2015. *J Am Coll Cardiol* **2017**, *70* (1), 1-25. DOI: 10.1016/j.jacc.2017.04.052  From NLM.
(2) World Health, O. Cardiovascular diseases (CVDs). **2025**. (acccessed 2025/09/12).
(3) Garcia, R.; Marijon, E.; Karam, N.; Narayanan, K.; Anselme, F.; Césari, O.; Champ-Rigot, L.; Manenti, V.; Martins, R.; Puymirat, E.; et al. Ventricular fibrillation in acute myocardial infarction: 20-year trends in the FAST-MI study. *Eur Heart J* **2022**, *43* (47), 4887-4896. DOI: 10.1093/eurheartj/ehac579  From NLM.
(4) Xu, X.; Wang, Z.; Yang, J.; Fan, X.; Yang, Y. Burden of cardiac arrhythmias in patients with acute myocardial infarction and their impact on hospitalization outcomes: insights from China acute myocardial infarction (CAMI) registry. *BMC Cardiovasc Disord* **2024**, *24* (1), 218. DOI: 10.1186/s12872-024-03889-w  From NLM.
(5) Demidova, M. M.; Holmqvist, F.; Erlinge, D.; Platonov, P. G. Ventricular arrhythmias during ST-segment elevation myocardial infarction and arrhythmic complications during recurrent ischaemic events. *Eur Heart J* **2024**, *45* (5), 393-395. DOI: 10.1093/eurheartj/ehad740   From NLM.

(6) Echivard, M.; Sellal, J. M.; Ziliox, C.; Marijon, E.; Bordachar, P.; Ploux, S.; Benali, K.; Marquié, C.; Docq, C.; Klug, D.; et al. Prognostic value of ventricular arrhythmia in early post-infarction left ventricular dysfunction: the French nationwide WICD-MI study. *Eur Heart J* **2024**, *45* (41), 4428-4442. DOI: 10.1093/eurheartj/ehae575   From NLM.

(7) Holmstrom, L.; Chugh, S. S. How to minimize in-hospital mortality from acute myocardial infarction: focus on primary prevention of ventricular fibrillation. *Eur Heart J* **2022**, *43* (47), 4897-4898. DOI: 10.1093/eurheartj/ehac578   From NLM.

(8) Wenzl, F. A.; Kraler, S.; Ambler, G.; Weston, C.; Herzog, S. A.; Räber, L.; Muller, O.; Camici, G. G.; Roffi, M.; Rickli, H.; et al. Sex-specific evaluation and redevelopment of the GRACE score in non-ST-segment elevation acute coronary syndromes in populations from the UK and Switzerland: a multinational analysis with external cohort validation. *Lancet* **2022**, *400* (10354), 744-756. DOI: 10.1016/s0140-6736(22)01483-0   From NLM.

(9) Yan, A. T.; Yan, R. T.; Tan, M.; Casanova, A.; Labinaz, M.; Sridhar, K.; Fitchett, D. H.; Langer, A.; Goodman, S. G. Risk scores for risk stratification in acute coronary syndromes: useful but simpler is not necessarily better. *Eur Heart J* **2007**, *28* (9), 1072-1078. DOI: 10.1093/eurheartj/ehm004   From NLM.

(10) Lewek, J.; Ptaszynski, P.; Klingenheben, T.; Cygankiewicz, I. The clinical value of T-wave alternans derived from Holter monitoring. *Europace* **2017**, *19* (4), 529-534. DOI: 10.1093/europace/euw292   From NLM.

(11) Verrier, R. L.; Huikuri, H. Tracking interlead heterogeneity of R- and T-wave morphology to disclose latent risk for sudden cardiac death. *Heart Rhythm* **2017**, *14* (10), 1466-1475. DOI: 10.1016/j.hrthm.2017.06.017   From NLM.

(12) Attia, Z. I.; Harmon, D. M.; Behr, E. R.; Friedman, P. A. Application of artificial intelligence to the electrocardiogram. *Eur Heart J* **2021**, *42* (46), 4717-4730. DOI: 10.1093/eurheartj/ehab649   From NLM.

(13) Siontis, K. C.; Noseworthy, P. A.; Attia, Z. I.; Friedman, P. A. Artificial intelligence-enhanced electrocardiography in cardiovascular disease management. *Nat Rev Cardiol* **2021**, *18* (7), 465-478. DOI: 10.1038/s41569-020-00503-2   From NLM.

(14) Li, J.; Aguirre, A. D.; Junior, V. M.; Jin, J.; Liu, C.; Zhong, L.; Sun, C.; Clifford, G.; Westover, M. B.; Hong, S. An Electrocardiogram Foundation Model Built on over 10 Million Recordings. *NEJM AI* **2025**, *2* (7), AIoa2401033. DOI: doi:10.1056/AIoa2401033.

(15) Fiorina, L.; Carbonati, T.; Narayanan, K.; Li, J.; Henry, C.; Singh, J. P.; Marijon, E. Near-term prediction of sustained ventricular arrhythmias applying artificial intelligence to single-lead ambulatory electrocardiogram. *Eur Heart J* **2025**, *46* (21), 1998-2008. DOI: 10.1093/eurheartj/ehaf073   From NLM.

(16) Kolk, M. Z. H.; Deb, B.; Ruipérez-Campillo, S.; Bhatia, N. K.; Clopton, P.; Wilde, A. A. M.; Narayan, S. M.; Knops, R. E.; Tjong, F. V. Y. Machine learning of electrophysiological signals for the prediction of ventricular arrhythmias: systematic review and examination of heterogeneity between studies. *EBioMedicine* **2023**, *89*, 104462. DOI: 10.1016/j.ebiom.2023.104462 From NLM.

(17) Rudin, C. Stop Explaining Black Box Machine Learning Models for High Stakes Decisions and Use Interpretable Models Instead. *Nat Mach Intell* **2019**, *1* (5), 206-215. DOI: 10.1038/s42256-019-0048-x From NLM.

(18) Trayanova, N. A.; Popescu, D. M.; Shade, J. K. Machine Learning in Arrhythmia and Electrophysiology. *Circ Res* **2021**, *128* (4), 544-566. DOI: 10.1161/circresaha.120.317872 From NLM.

(19) Khan, W.; Leem, S.; See, K. B.; Wong, J. K.; Zhang, S.; Fang, R. A Comprehensive Survey of Foundation Models in Medicine. *IEEE Rev Biomed Eng* **2025**, *Pp*. DOI: 10.1109/rbme.2025.3531360 From NLM.

(20) Bommasani, R. On the opportunities and risks of foundation models. *arXiv preprint arXiv:2108.07258* **2021**.

(21) Chen, R. J.; Ding, T.; Lu, M. Y.; Williamson, D. F. K.; Jaume, G.; Song, A. H.; Chen, B.; Zhang, A.; Shao, D.; Shaban, M.; et al. Towards a general-purpose foundation model for computational pathology. *Nat Med* **2024**, *30* (3), 850-862. DOI: 10.1038/s41591-024-02857-3 From NLM.

(22) Barceló, P.; Kozachinskiy, A.; Romero, M.; Subsercaseaux, B.; Verschae, J. Explaining k-Nearest Neighbors: Abductive and Counterfactual Explanations. *Proceedings of the ACM on Management of Data* **2025**, *3*, 1-26. DOI: 10.1145/3725234.

(23) Salehinejad, H.; Sankar, S.; Barfett, J.; Colak, E.; Valaee, S. Recent advances in recurrent neural networks. *arXiv preprint arXiv:1801.01078* **2017**.